\documentclass[conference]{IEEEtran}
\IEEEoverridecommandlockouts
\usepackage{cite}
\usepackage{amsmath,amssymb,amsfonts}
\usepackage{graphicx}
\usepackage{textcomp}
\usepackage{xcolor}
\usepackage{hyperref}
\hypersetup{colorlinks,allcolors=black}

\usepackage{multirow}
\usepackage[linesnumbered,ruled,vlined]{algorithm2e}
\usepackage{algcompatible}
\graphicspath{ {img/} }
\usepackage{caption}
\usepackage{subcaption}
\usepackage{bbding}

\def\BibTeX{{\rm B\kern-.05em{\sc i\kern-.025em b}\kern-.08em
    T\kern-.1667em\lower.7ex\hbox{E}\kern-.125emX}}
\begin{document}
\pdfoutput=1
\title{FisheyePP4AV: A privacy-preserving method for autonomous vehicles on fisheye camera images\\
}

\author{\IEEEauthorblockN{Linh Trinh\textsuperscript{\textdagger}\textsuperscript{\Envelope}}
\IEEEauthorblockA{\textit{Data Science Department} \\
\textit{VinFast}\\
Hanoi, Vietnam \\
linhtk.dhbk@gmail.com}
\and
\IEEEauthorblockN{Bach Ha\textsuperscript{\textdagger}}
\IEEEauthorblockA{\textit{Falculty of Applied Mathematics and Informatics} \\
\textit{Hanoi University of Science and Technology}\\
Hanoi, Vietnam \\
hasybach1103@gmail.com}
\and
\IEEEauthorblockN{Anh Tu Tran}
\IEEEauthorblockA{\textit{Falculty of Information Security} \\
\textit{Academy of Cryptography Techniques}\\
Hanoi, Vietnam \\
tutran@actvn.edu.vn}
\thanks{\textsuperscript{\textdagger} Equally contributed.}
\thanks{\textsuperscript{\Envelope} Corresponding author.}
}

\maketitle

\begin{abstract}
In many parts of the world, the use of vast amounts of data collected on public roadways for autonomous driving has increased. In order to detect and anonymize pedestrian faces and nearby car license plates in actual road-driving scenarios, there is an urgent need for effective solutions. As more data is collected, privacy concerns regarding it increase, including but not limited to pedestrian faces and surrounding vehicle license plates. Normal and fisheye cameras are the two common camera types that are typically mounted on collection vehicles. With complex camera distortion models, fisheye camera images were deformed in contrast to regular images. It causes computer vision tasks to perform poorly when using numerous deep learning models. In this work, we pay particular attention to protecting privacy while yet adhering to several laws for fisheye camera photos taken by driverless vehicles. First, we suggest a framework for extracting face and plate identification knowledge from several teacher models. Our second suggestion is to transform both the image and the label from a regular image to fisheye-like data using a varied and realistic fisheye transformation. Finally, we run a test using the open-source PP4AV dataset.  The experimental findings demonstrated that our model outperformed baseline methods when trained on data from autonomous vehicles, even when the data were softly labeled. The implementation code is available at our github: \href{https://github.com/khaclinh/FisheyePP4AV}{https://github.com/khaclinh/FisheyePP4AV}.
\end{abstract}


\section{Motivation}\label{sec:intro}
Data privacy protection for autonomous vehicles is turning into a serious issue that requires attention. Companies and research teams have started gathering a lot of data as machine learning has been employed more and more in autonomous driving for development and validation. Since 2018~\cite{Waymo5mil}, Waymo has accumulated 5 million miles. Only in 2020 \cite{Cruise770} did Cruise collect more than 770,000 miles. With more data being collected comes more accountability for data privacy. For instance, laws from the European GDPR \cite{gdprEU}, California CCPA \cite{ccpaCali}, Chinese CSL \cite{cslChina}, or Japanese APPI \cite{appiJapan} must be followed when collecting data on public highways. According to the regulations, participants' personal identity information must be protected and deleted upon request. Numerous commercial devices that de-identify acquired data have been released in response to these regulations, usually by obscuring camera data. The faces and license plates are made anonymous using Brighter AI\footnote{https://brighter.ai/video-redaction-in-automotive/}, Facebook Mapillary\footnote{https://www.mapillary.com/geospatial}, or UAI Anonymizer \cite{uaiAnonymizer}. Celantur\footnote{https://www.celantur.com/} goes even further by masking people's faces, license plates, bodies, even entire automobiles.

To the best of our knowledge, \cite{pp4av} is the first open benchmarking dataset used to assess a model on autonomous driving that protects privacy. 3,447 driving photos with faces and license plates on both fisheye camera and regular camera photographs are included in this dataset. In order to show the limitations of those models on the domain of autonomous driving, they also supplied a based line model and a thorough comparison with several pretrained models. Fisheye camera images are rarely utilized for training and testing a privacy-preserving model, in contrast to regular photos. Since the majority of previously trained models were trained on commonplace photos, they typically continue to perform admirably on new domain datasets. Due to these models' poor performance on fisheye photos, a unique technique is needed to help them adapt to the data from fisheye cameras.

In this article, we provide a new method for face and license plate recognition in fisheye camera images used for autonomous driving. We developed a collection of varied and realistic distortion methods to change the data from normal to fisheye-like in order to address the low performance on fisheye camera images. To be more precise, we employ four distortion models developed by \cite{fisheye@2006, fisheyeRadial, circileFisheye} for producing a variety of fisheye-like training data. In order to overcome the shortcomings of the lack of ground truth for training, we extend the baseline model put forward by citepp4av to present an enhanced framework for training data that draws expertise from numerous teachers.

In summary, the main contributions of this work are in 3 folds:
\begin{itemize}
    \item We propose our framework for training a model for face and license plate anonymization via model distillation from several teachers.
    \item We propose a fisheye transformation to convert both the image and the pseudo label supplied by the teacher model into fisheye-like data for training the student model. For improved adaption, this fisheye transformation includes a variety of realistic distortion kinds.
    \item We train our anonymization model for self-driving cars. Despite the fact that our model was trained without any actual annotated dataset, the experimental results indicated that it outperformed the baseline model from \cite{pp4av}.
\end{itemize}

The remainder of this paper is organized as follow: we present a our method in the section \ref{section:method}. The next section \ref{section:exp} we present experiment and it's result to show the performance of our proposed method. Finally, section \ref{section:conclusion} aim to conclude our paper.

\section{Methods}\label{section:method}
\begin{figure*}[h]
\begin{center}
    \includegraphics[width=0.9\linewidth]{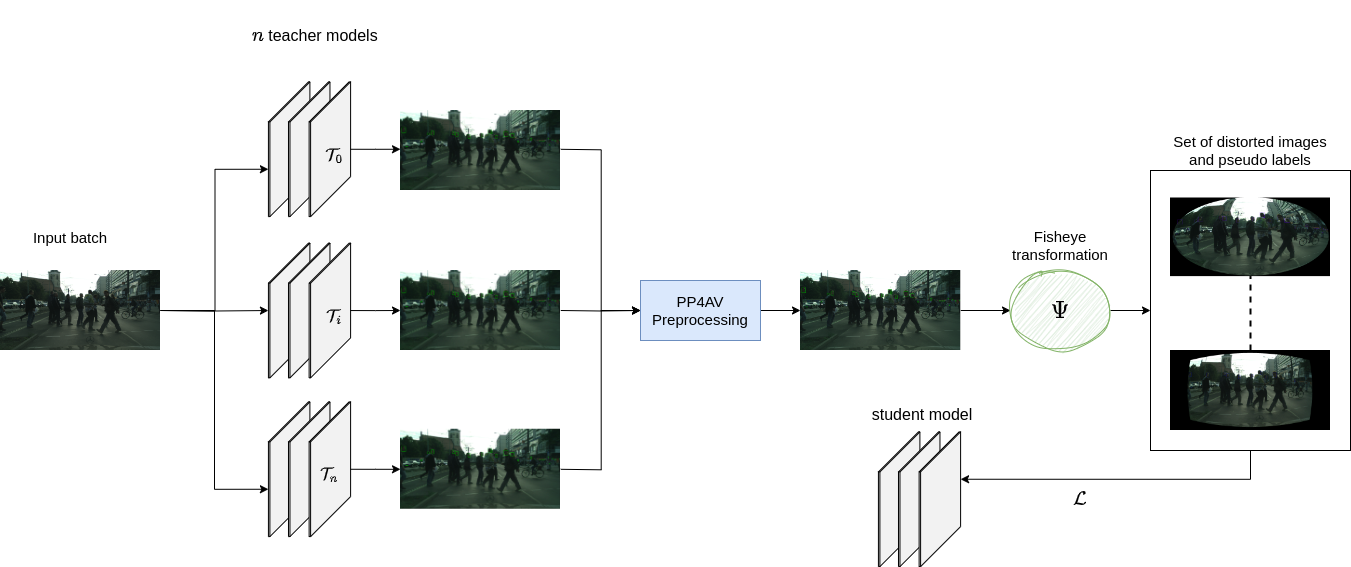}
\end{center}
   \caption{Illustration of our framework. Due to lack of ground truth, we leverage multiple models trained on other task for face and plate detection as the teacher models which are used to distill knowdlege to our model via pseudo label generation. Fisheye transformation $\Psi$ is used to transform pseudo labels and images into fisheye-like pseudo labels and images.}
\label{fig:fw}
\end{figure*}

Our framework is illustrated in the Figure \ref{fig:fw}. The main parts of the framework are the multiple teacher models, the PP4AV preprocessing, the fisheye transformation, and our student model. Due to the lack of ground truth, we use multiple models trained on other tasks to teach our model how to detect faces and license plates. This is done through pseudo label generation. After generating pseudo labels for training batch data, we use pseudo label preprocessing from PP4AV to aggregate the pseudo labels and confidence scores from various teacher models into a single set. This step improved the power of using multiple models to make high-quality, confident pseudo labels that will be used for training in the next step. The fisheye transformation will turn both images and their pseudo labels into data that looks like a fisheye. Lastly, the fisheye-like data we got at the end are used to train our model.

\subsection{Teacher and student models.}
Similar as PP4AV baseline model, we select UAI Anonymizer, YOLO5Face \cite{yolo5face2021}, and RetinaFace \cite{retinaface_deng2020} as three teacher models for face detection, and UAI Anonymizer as teacher model for license plate detection.

For student model, we keep the same setting of modification of YOLOX \cite{yolox2021} which presented in PP4AV \cite{pp4av} for our student model. Three changes are made: (1) the Focus layer is swapped out for a stem block structure; (2) the SSP block is modified to use a smaller kernel; and (3) a P6 output block with a stride of 64 is added.

\subsection{Fisheye transformation $\Psi$}
We define fisheye image transformation $\Psi$ as a function of a set distortion transformation, i.e. $\Psi=\mathcal{F}(\phi_1, \phi_2,...,\phi_n)$ where $\phi_i$ is $i$th distortion transformation function which transform normal data to fisheye-like data by distortion function, $\mathcal{F}$ is an aggregation function which is used to aggregate the result from these distortion transformations. $\mathcal{F}$ can be a discrete or continuous function. In our work, to most simplify, we setup $\mathcal{F}$ as a random selection from a set of function $\phi$. For more detail, given a normal data $x$, the output of fisheye transformation is:
\begin{equation}
    x'=\Psi(x)=\mathcal{F}_{\phi_1,\phi_2,...,\phi_n}(x)=\phi_i(x)
\end{equation}
where $i$ is the randomly selected distortion function.

In this work, we apply four distortion functions $\phi$ for transforming normal data to fisheye-like data which usually applied to autonomous vehicle's camera data. Four functions are circular \cite{circileFisheye}, rectangular \cite{fisheye@2006}, radial \cite{fisheyeRadial} and tangential \cite{fisheyeRadial} transformation functions.
The center of the square data is at $(0, 0)$, and the coordinates for the four corners are $(1, 1)$. Let indicate $(x, y)$ be the normalized coordinates of the input. By translating to the optical center and dividing by the focal length in pixels, one can determine the normalized image coordinates from the pixel coordinates. $(x_d, y_d)$ are used to indicate the deformed points.

\textbf{Circular transformation.}
The conversion of a normal patch to a circular batch is expressed in the equations below. The circular patch formed is referenced by the output coordinates $(x',y')$.
\begin{equation}\label{eq:circular1}
    \left( \begin{array}{cc}
    x' \\
    y'\\
    \end{array}\right)=\left( \begin{array}{cc}
    x\cdot \sqrt{1-\frac{y^2}{2}} \\
    y\cdot \sqrt{1-\frac{x^2}{2}} \\
    \end{array}\right)
\end{equation}
Equation \ref{eq:circular2} furthers squeezes the circular image towards the perimeter. Here $r=\sqrt{(x')^2+(y')^2}$ is the radial distance from the center of the circular path.
\begin{equation}\label{eq:circular2}
    \left( \begin{array}{cc}
    x_d \\
    y_d\\
    \end{array}\right)=\left( \begin{array}{cc}
        x'\cdot e^{-\frac{r^2}{4}} \\
        y'\cdot e^{-\frac{r^2}{4}} \\
    \end{array}\right)
\end{equation}

\textbf{Rectangular transformation.} Rectangular transformation function are expressed as below equation:
\begin{equation}\label{eq:circular2}
    \left( \begin{array}{cc}
    x_d \\
    y_d\\
    \end{array}\right)=r_f\left( \begin{array}{cc}
        x\sin{\gamma} \\
        y\cos{\gamma} \\
    \end{array}\right)
\end{equation}
where $r_f$ is determined as below equation:
\begin{equation}
    r=f\tan (\frac{r_f}{f})
\end{equation}
and $\gamma$ is defined as below equation:
\begin{equation}
    \gamma = \arctan \frac{x}{y}
\end{equation}
which $r=\sqrt{x^2+y^2}$, $f$ is the focal length.

\textbf{Radial transformation.}
Radial transformation function are expressed as below equation:
\begin{equation}
    \left( \begin{array}{cc}
    x_d \\
    y_d\\
    \end{array}\right)=\left( \begin{array}{cc}
        x\left (1+k_1\cdot r^2+k_2\cdot r^4+k_3\cdot r^6\right ) \\
        y\left (1+k_1\cdot r^2+k_2\cdot r^4+k_3\cdot r^6\right ) \\
    \end{array}\right)
\end{equation}
where $r=\sqrt{x^2+y^2}$, and $k_1, k_2, k_3$ are radial distortion coefficients of the lens.

\textbf{Tangential transformation.} Tangential transformation function are expressed as below equation:
\begin{equation}
    \left( \begin{array}{cc}
        x_d \\
        y_d\\
        \end{array}\right)=\left( \begin{array}{cc}
            x+\left [ 2\cdot p_1\cdot x\cdot y +p_2\cdot (r^2 + 2\cdot x^2) \right ] \\
            y+\left [ p_1\cdot (r^2+2\cdot y^2)+2\cdot p_2 \cdot x \cdot y \right ] \\
        \end{array}\right)
\end{equation}
where $p_1, p_2$ are tangential distortion coefficients of the lens.

\subsection{Loss function}
Similar to PP4AV baseline model \cite{pp4av}, the loss function is as follows:
\begin{equation}
    \mathcal{L} = \lambda \cdot \mathcal{L}_{iou} + \mathcal{L}^{fl}_{cls} + \mathcal{L}^{fl}_{obj} + \gamma \cdot \mathcal{L}_{KL}
\end{equation}
where $\mathcal{L}^{fl}_{cls}$, $\mathcal{L}^{fl}_{obj}$ are focal losses for classification and regression, respectively, and $\gamma$ is the weight factor for KL divergence loss $\mathcal{L}_{KL}$, $\lambda$ is the weight factor for IoU loss $\mathcal{L}_{iou}$.

\section{Experiments}\label{section:exp}
\subsection{Settings}
In this section, we present experiment and the performance result of our model.

\textbf{Datasets.}
We build our model training dataset from open datasets for autonomous driving that are already available. Although these datasets cover a wide range of contexts, they lack face and license plate annotations, which are disadvantages for our goal. We ignore all general-purpose public datasets because they are unrelated to the driving situation because we are focusing on public datasets for self-driving vehicles. Another issue is that no facial or license plate annotations can be found in any of the available datasets for self-driving automobiles.  In our method, we seek to leverage the pretrained model (which we subsequently use as a teacher model to train our model) to teach our model via prediction rather than annotating and feeding the prediction of these models into our model, as we have previously researched. The training and validation sets in this experiment are summarized in table \ref{table:trainingDatasetTable}. For training, 62,927 photos from six public datasets are combined, while 10,250 images are used for validation.
\begin{table}[ht]
    \begin{center}
    \begin{tabular}{|l|c|c|c|c|}
    \hline
    Dataset           & Resolution & Train & Val  \\ \hline
    Cityscape~\cite{cityscape_cordts2016}            & 2,048$\times$1,024  & 2,921  & 488  \\ \hline
    BDD100K~\cite{bdd100k_yu2020}          & 1,280$\times$720   & 41,568 & 7,370 \\ \hline
    Comma2K19~\cite{comma2k19_1812.05752}       & 1,164$\times$874   & 6,358  & 1,414 \\ \hline
    Bosch~\cite{bosch_boxy_behrendt2019}               & 2,464$\times$2,056  & 3,500  & 750  \\ \hline
    LeddarPixSet~\cite{leddar_pixset_deziel2021}       & 1,440$\times$1,080  & 1,062  & 228  \\ \hline
    Kitti~\cite{kitti_geiger2013}          & 1,240$\times$376   & 7,518  & 0   \\ \hline
             &    \textbf{Total}  & 62,927  & 10,250   \\ \hline
    \end{tabular}
    \end{center}
    \caption{The overview and number of images in the training and validation set of the model.}
    \label{table:trainingDatasetTable}
\end{table}
We use the fisheye data and it's annotation from \cite{pp4av} for evaluation. This data consists of 244 fisheye camera images which originally provided by WoodScape \cite{woodscape_yogamani2019} are well annotated both face and license plate.

\textbf{Pseudo label transformation.} Each label was altered to the new coordinate system for pseudo label processing. Four corner points and four edge midpoints were chosen for each bounding box, totaling eight points. These eight points were mapped to the coordinate system of the fisheye image using the same transformation function as the associated image. In the coordinate system of the fisheye image, the new points constitute a polygon. These additional eight points were used to find the smallest axis-aligned bounding rectangle, which was then saved as the new bounding-box label for the fisheye image.

\textbf{Experiment setting.}
For evaluation metrics, we use standard metrics AP50 and AR50 which usually used for object detection to measure average precision and average recall at IOU=0.5.
For fisheye transformation, we set $p_1,p_2$ are 0.2, 0.1 respectively for tangential distortion, $k_1,k_2,k_3,k_4$ are 0.2, 0.1, 0.05, 0.05 for radial distortion, and focal length is 250 for rectangular distortion.
As a preprocessing step, data augmentation is used to increase the training's robustness. In particular, there is a 50\% probability of doing the following augmentations: horizontal flips, brightness adjustment (0.2), saturation adjustment (0.2), contrast adjustment (0.2), hue jitter (0.1), mosaic, rotation, and shear.
We train our model with hyperparameters setting: batch size 32, image resize to 640x640, learning rate is 0.0001, and we use SGD optimizer for training.
All experiments were conducted on an NVIDIA DGX A100 server with 4 GPUs.

\subsection{Quantitative results}
The Table \ref{table:performanceResultTable} show the comparison of our model with PP4AV \cite{pp4av} on AP50 and AR50 for face and plate objects individually. The results show that our model outperform the baseline PP4AV in both AP50 and AR50. For face detection, our model improves 1.89\% and 1.21\% on AP50 and AR50 separately. For license plate detection, the improvement is minor at AP50 with 0.24\% increasing and 1.94\% increasing at AR50. The promising results show the significant improvement of our method which is realistic and adaptive to specific domain such as fisheye camera data in autonomous vehicles.
\begin{table}[ht]
    \begin{center}
    \begin{tabular}{|c|c|c|c|} \hline
    \multirow{2}{*}{}           & \multicolumn{1}{c}{\multirow{2}{*}{Methods}} & \multicolumn{2}{|c|}{Metrics} \\ \cline{3-4}
            \multicolumn{1}{|l|}{}           & \multicolumn{1}{c|}{}                               & AP50           & AR50          \\ \hline \hline
    \multirow{2}{*}{\rotatebox[origin=c]{90}{Face}}
                                   & PP4AV \cite{pp4av}                              & 59.2\%   & 63.92\%  \\ \cline{2-4}
                                   & \textbf{Our}                                & \textbf{61.09}\%   & \textbf{65.13}\%  \\ \hline \hline
    \multirow{2}{*}{\rotatebox[origin=c]{90}{Plate}} 
                                   & PP4AV \cite{pp4av}                           & 49.53\%   & 58.17\%  \\ \cline{2-4}
                                   & \textbf{Our}                         & \textbf{49.77}\%   & \textbf{60.11\%}  \\ \hline
    \end{tabular}
    
    \end{center}
    \caption{Comparison of performance on fisheye test set of PP4AV \cite{pp4av} on Average Precision (AP) and Average Recall (AR) scores.}
    \label{table:performanceResultTable}
\end{table}

\subsection{Qualitative analysis}
\begin{figure*}[ht]
\captionsetup[subfigure]{labelformat=empty}
\setlength\tabcolsep{3pt} 
\centering  
\begin{tabular}{l c c}
\begin{subfigure}{0.05\linewidth}\caption{\rotatebox[origin=c]{90}{Ground truth}} \end{subfigure} 
  & \includegraphics[width=0.4\linewidth]{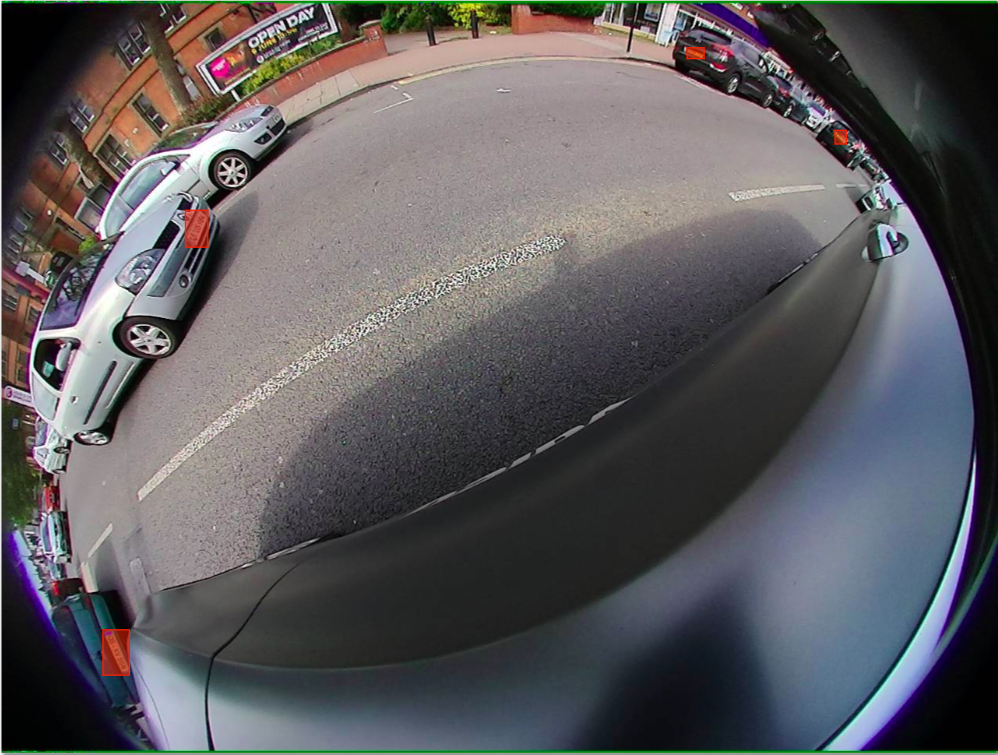} 
  & \includegraphics[width=0.4\linewidth]{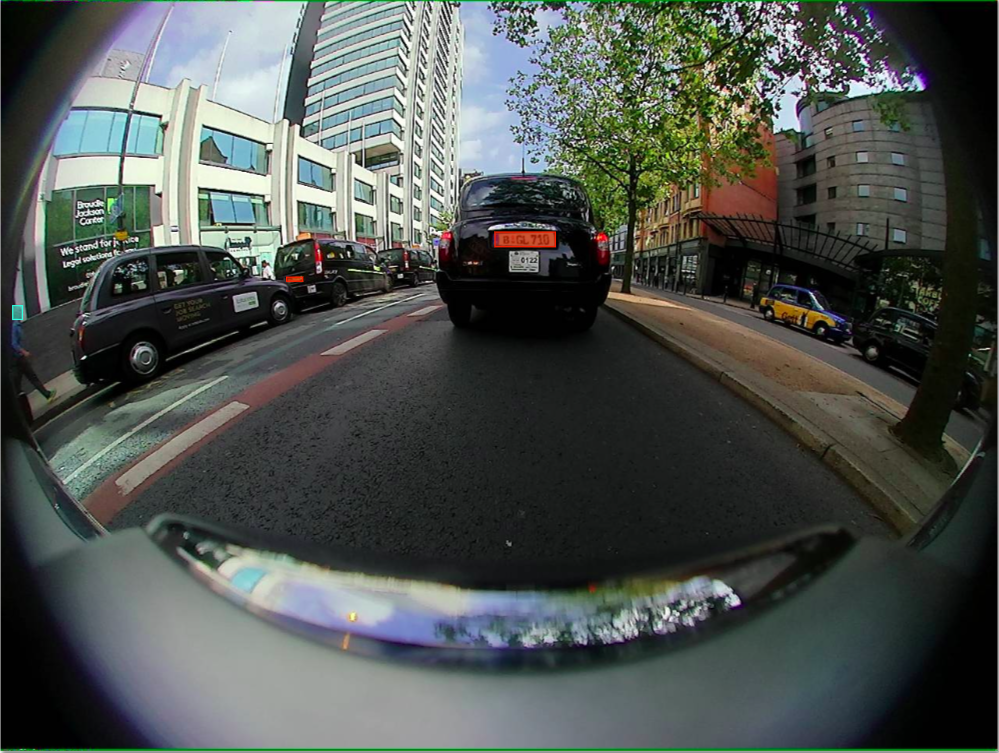} \\
 
 \begin{subfigure}{0.05\linewidth}\caption{\rotatebox[origin=c]{90}{PP4AV}} \end{subfigure} 
  & \includegraphics[width=0.4\linewidth]{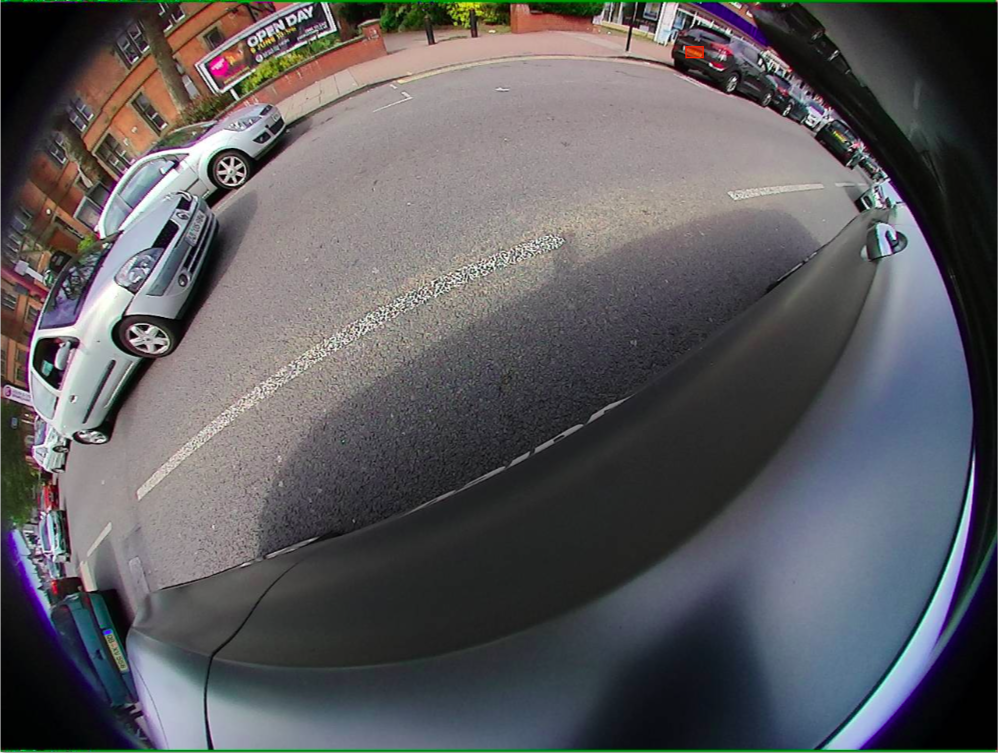} 
  & \includegraphics[width=0.4\linewidth]{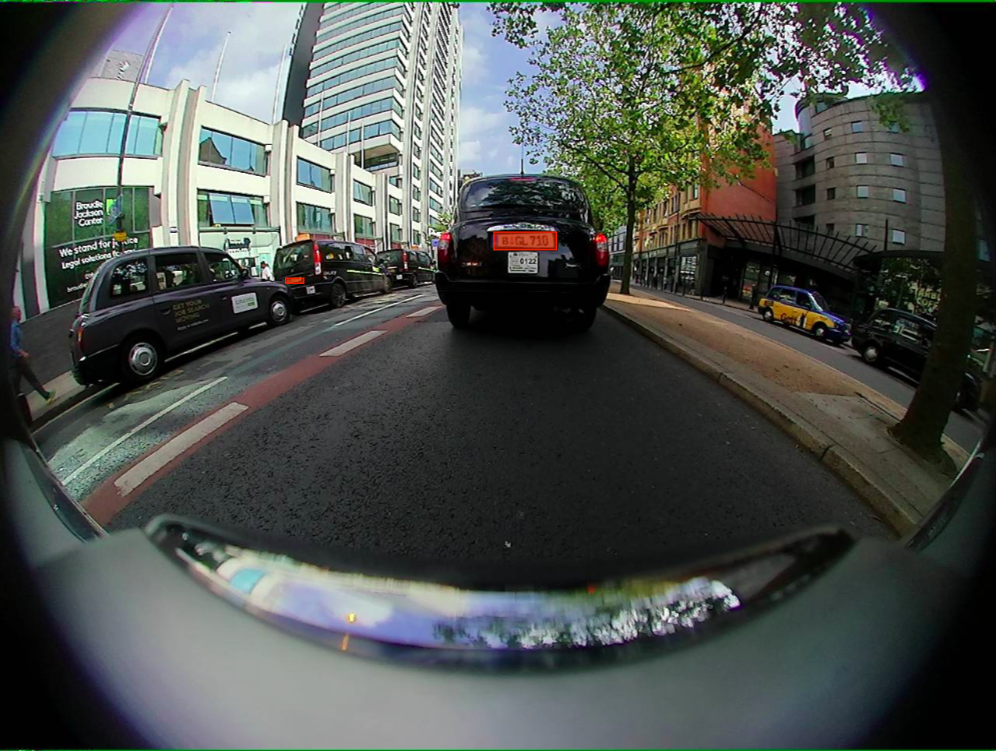} \\
 
 \begin{subfigure}{0.05\linewidth}\caption{\rotatebox[origin=c]{90}{\textbf{Ours.}}} \end{subfigure} 
  & \includegraphics[width=0.4\linewidth]{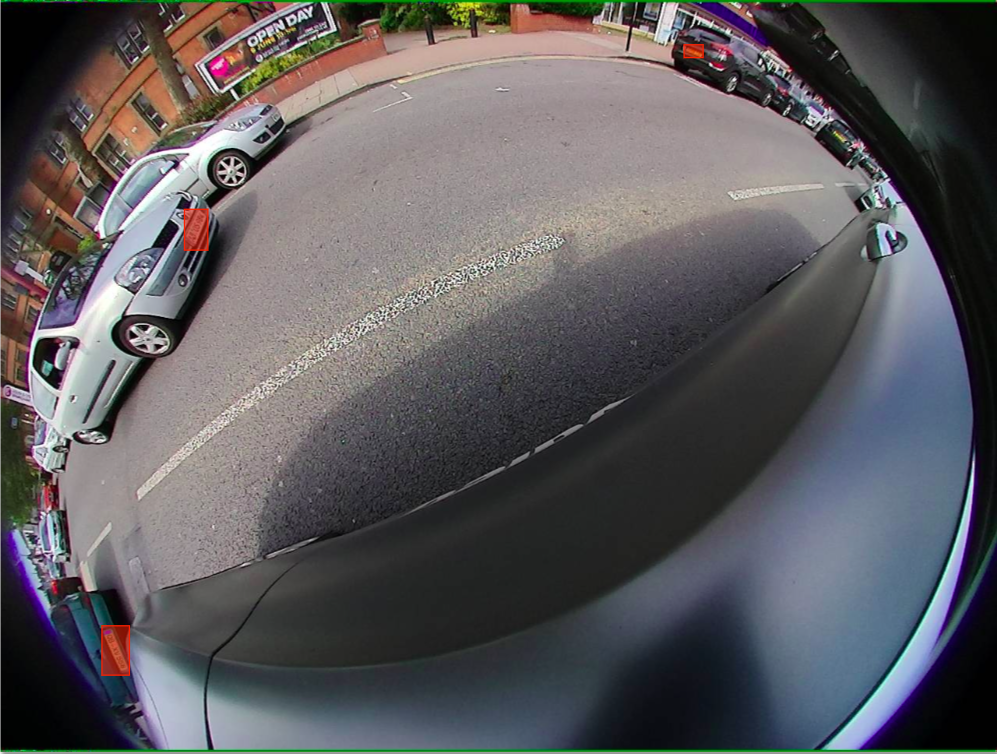} 
  & \includegraphics[width=0.4\linewidth]{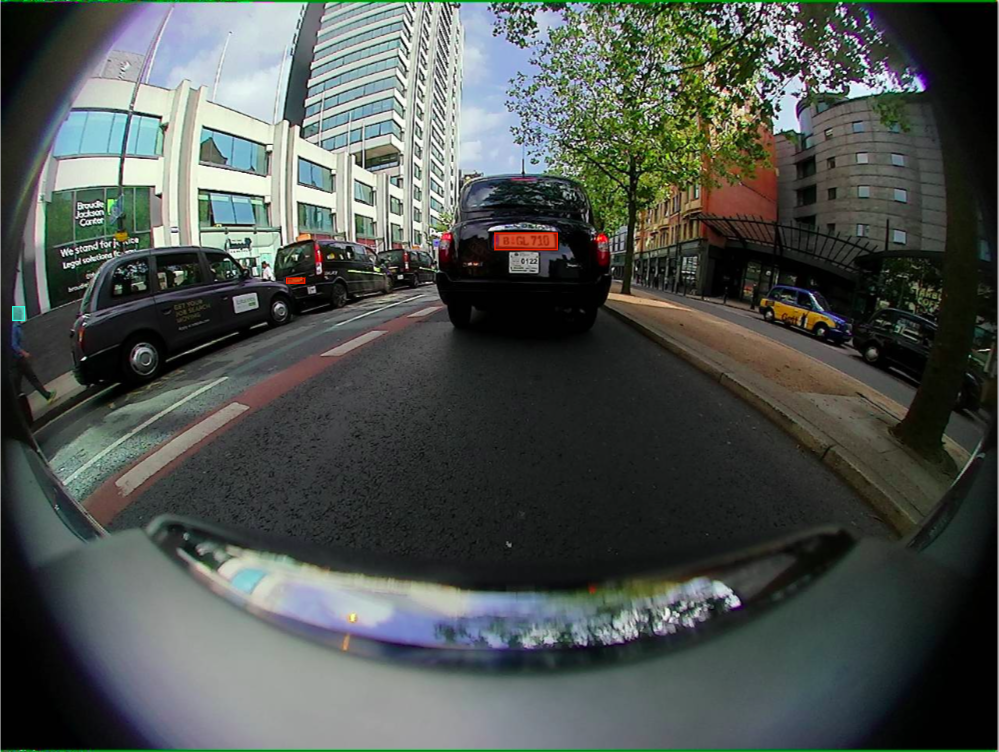} \\

\end{tabular}
\caption{Example results for face and license plate detection on a fisheye camera image from the \cite{pp4av} dataset.}
\label{fig:qualitative}
\end{figure*}

For qualitative analysis, we randomly select some images from the test set. Figure \ref{fig:qualitative} compares our model to the baseline model from \cite{pp4av} and the ground truth. Because the first image was highly distorted, the \cite{pp4av} model failed to recognize almost all of the plates. Our algorithm, which was trained on fisheye-like data, can detect the majority of clear deformed plates. In comparison to the real world, our model missed only one far and not transparent plate. \cite{pp4av} has a misdetection on a human face on the boundary of a fisheye picture in the right shot. When the human face is positioned on the boundary of a significantly distorted image, as in the sample, the face's shape is also strongly warped. As in the real world, our model can recognize it correctly. The results in the sampled data show that the model trained on just normal images may not recognise distorted objects in fisheye images. It demonstrates the need of modifying and training models on similar types of data, such as fisheye-like images, for improved adaptability to fisheye camera images in autonomous driving.

\section{Conclusions}\label{section:conclusion}
In this paper, we present a method for data anonymization on fisheye camera images of autonomous driving to guarantee regulations such as EU GDPR, CCPA, and CSL. Due to a lack of ground truth for training, we propose a framework to leverage knowledge from multiple teacher models trained on other tasks for face and license plate detection for training our model by distilling information via autonomous vehicle data. Furthermore, we propose a fisheye transformation that transforms normal data, such as an image and its label, into fisheye-like data by applying diverse and realistic distortion functions that are usually used for autonomous vehicle data. The experiment results on the fisheye test set data of PP4AV show our model improves significantly in performance compared to the baseline model. This promising result shows that our proposed method adapts efficiently to fisheye camera image-based object detection. In future work, we will consider extending our framework for computer vision tasks to other domain data based on fisheye camera data.

\bibliographystyle{IEEEtran}

\bibliography{reference.bib}

\end{document}